# A Typology of Data Anomalies

Ralph Foorthuis

UWV, La Guardiaweg 116, 1040 HG Amsterdam, The Netherlands
`ralph.foorthuis@uwv.nl`

**Abstract.** Anomalies are cases that are in some way unusual and do not appear to fit the general patterns present in the dataset. Several conceptualizations exist to distinguish between different types of anomalies. However, these are either too specific to be generally applicable or so abstract that they neither provide concrete insight into the nature of anomaly types nor facilitate the functional evaluation of anomaly detection algorithms. With the recent criticism on 'black box' algorithms and analytics it has become clear that this is an undesirable situation. This paper therefore introduces a general typology of anomalies that offers a clear and tangible definition of the different types of anomalies in datasets. The typology also facilitates the evaluation of the functional capabilities of anomaly detection algorithms and as a framework assists in analyzing the conceptual levels of data, patterns and anomalies. Finally, it serves as an analytical tool for studying anomaly types from other typologies.

Classification · Data mining · Exploratory analytics · Pattern recognition
Machine learning

## 1 Introduction

*Anomalies* are cases that are in some way unusual and do not appear to fit the general patterns present in the dataset [1, 2, 3]. Such cases are often also referred to as *outliers*, *novelties* or *deviant observations* [3, 4]. Anomaly detection (AD) is the process of analyzing the dataset to identify these deviant cases. Anomaly detection can be used for various goals, such as fraud detection, data quality analysis, security scanning, process and system monitoring, and data cleansing prior to training statistical models [1, 2, 3, 4].

Several ways to distinguish between different kinds of anomalies have been presented in the literature. These conceptualizations, however, are either only relevant for specific situations or too abstract to provide a clear and concrete understanding of anomalies (see sections 2 and 4). This paper therefore presents a *typology of anomalies* that offers a theoretical and tangible *understanding* of the nature of different types of anomalies, assists researchers with *evaluating* the functional capabilities of their anomaly detection algorithms, and as a framework aids in *analyzing*, i.a., the conceptual levels of data and anomalies. A preliminary version has been presented briefly in [1, 5] to evaluate an unsupervised non-parametric AD algorithm. This paper extends that initial typology and discusses its theoretical properties in more depth.

A clear understanding of the types of anomalies that can be encountered in datasets is relevant for several reasons. First, it is important in statistics, data science, machine learning, analytics and knowledge-based systems to have a fundamental and tangible understanding of anomalies, of the various anomaly types that exist, and of their defining characteristics. In this context, the typology presented here not only helps in theoretically understanding the nature of data and (deviations from) patterns, but also provides a functional evaluation framework that enables researchers to demonstrate which anomaly type(s) their AD algorithms are able to detect. Second, with the recent criticism on 'opaque' and 'black box' analytics methods that may result in unfair outcomes [6, 7], it has become clear that it is undesirable to have algorithms and analysis results that lack transparency and cannot be interpreted meaningfully. This is especially true for AD algorithms, as these may be used to identify and act on 'suspicious' cases. Although the typology presented here does not make the algorithms themselves more transparent, a clear understanding of (the types of) anomalies and their properties helps in making the results of data analyses understandable and transparent. Third, even if statistical and machine learning algorithms are functionally transparent and understandable, the implementations of these algorithms – and the knowledge-based systems they are part of – may be done poorly or simply fail due to overly complex real-world settings [8, 9]. The results of data analyses conducted in practice may thus prove to be incorrect and unpredictable. A deep understanding of anomalies is therefore needed to determine whether detected cases indeed constitute true anomalies. This is especially relevant for unsupervised AD algorithms, as these are often not used with known labelled data. Finally, the *no free lunch* theorem, which posits that no single algorithm will show superior performance in all problem domains, also holds for anomaly detection [10, 11, 12]. Individual AD algorithms are generally not able to detect all types of anomalies and will differ in their performance. In addition, more complex algorithms do not necessarily perform better than relatively simple ones. The typology assists researchers in making transparent which algorithms are able to detect what types of anomalies to what degree.

This paper proceeds as follows. Section 2 discusses related research. Section 3 presents the typology of anomalies. Section 4 discusses the properties of the typology and compares it with other research. Finally, section 5 is for conclusions.

**Fig. 1.** Overview of typologies and anomaly types (G/S refers to General vs. Specific)

| Ref. | G/S | Anomaly types | Use of Explicit Dimensions |
|---|---|---|---|
| [1,5,18] | G | Extreme value anomaly, Multidimensional numerical anomaly, Sparse class anomaly, Multidimensional mixed data anomaly | Nature of data (c.f. Types of data), Number of interacting attributes (c.f. the Cardinality of relationship) |
| [3] | G | Weak outlier, Strong outlier | None (random noise is a candidate) |
| [13] | G | Point anomaly, Contextual anomaly, Collective anomaly | None |
| [2, 17] | S | Outliers, High-leverage points, Influential points | None |
| [15] | S | Additive outlier, Transitory change outlier, Level shift outlier, Innovational outlier | None |
| [16] | S | Isolated outliers, Shift outliers, Amplitude outliers, Shape outliers | None |

## 2    Related work

The literature acknowledges various ways to distinguish between types of anomalies. In [3] a distinction is made between a *weak outlier* (noise that can be attributed to the statistical variation of a random variable) and a *strong outlier* (a true anomaly that may be generated by a mechanism different from the one generating the normal cases). The general typology presented in [13] differentiates between three types. A *point anomaly* refers to one or several individual cases that are deviant with respect to the rest of the data. A *contextual anomaly* appears normal at first sight, but is deviant when an explicitly mentioned context is taken into account [also see 14]. For example, when a measured temperature value is not low in general, but only in the summer season. Finally, a *collective outlier* refers to a group of data points that belong together and, as a group, deviates from the rest of the data. This requires 'dependent data', in which individual cases (rows) are related by e.g. a time, space or identification attribute. A unique sequence of events is an example of such a collective outlier.

Several more specific and concrete typologies are also known, especially from time series or sequence analysis. In [15] several within-sequence types are presented. An *additive outlier* in this context is an isolated spike during a short period, whereas a *transitory change outlier* is a spike that requires some time to disappear. A *level shift outlier* is a sudden but structural change to a higher or lower value level, whereas an *innovational outlier* may show shifts in both the trend and the seasonal pattern. The taxonomy presented in [16] focuses on between-sequences anomalies and makes a distinction between *isolated outliers*, *shift outliers, amplitude outliers* and *shape outliers*. Even more types can be acknowledged, such as *deviant cycle anomalies*. Fig. 5 in the Discussion section illustrates several of these types. Another example of a specific typology is known from regression analysis, in which it is common to distinguish between *outliers*, *high-leverage points* and *influential points* [2, 17].

The above mentioned typologies, summarized in Fig. 1, are either too general and too abstract to provide a clear and concrete understanding of anomaly types, or feature well-defined types that are only relevant for a specific purpose (such as time series analysis or regression modeling). This paper therefore presents a typology of anomalies that is general but still offers a meaningful, tangible and useful description of the various types of anomalies. Such a typology has significant value for both practitioners and researchers. It seems that a similar typology, grounded in the fundamental dimensions of data types and attribute relationships, has not been published before (note that many typologies regarding anomaly detection *techniques* do exist).

## 3    Typology of Anomalies

This section presents the general typology of anomalies that offers a theoretical, detailed and concrete understanding of the types of anomalies that can be encountered in datasets. It also gives researchers a tool to evaluate which types of anomalies can be detected by a given AD algorithm, assists in analyzing the conceptual levels of patterns and anomalies, and aids in studying anomaly types from other typologies.

The typology uses two dimensions, each of which describes a fundamental aspect of the nature of data, to distinguish between anomaly types. The first dimension represents the **types of data** involved in describing the behavior of the cases. This refers to the data types of the attributes (i.e. variables) that are involved in the anomalous character of a deviant case and thus have to be handled appropriately during the analysis in order for it to be detected. The data types are:

- *Continuous*: The variables that capture the anomalous behavior are all numeric in nature. Examples of such variables are age, height and temperature.
- *Categorical*: The variables that capture the anomalous behavior all represent codes or class values. This includes binomial and multinomial attributes. Examples of such variables are gender, country, color and animal species.
- *Mixed*: The variables that capture the anomalous behavior are both continuous and categorical in nature. At least one attribute of each type is present.

Not all data types acknowledged in so-called multimodal objects are included here. The reason is that e.g. video, audio and free text anomalies can generally be reduced to class- or number-based deviations, or require a very specific analysis.

|  |  | **Types of Data** | | |
|---|---|---|---|---|
|  |  | **Continuous attributes** | **Categorical attributes** | **Mixed attributes** |
| **Cardinality of Relationship** | **Univariate** Described by individual attributes (independence) | Type I Extreme value anomaly | Type II Rare class anomaly | Type III Simple mixed data anomaly |
|  | **Multivariate** Described by multi-dimensionality (dependence) | Type IV Multidimensional numerical anomaly | Type V Multidimensional rare class anomaly | Type VI Multidimensional mixed data anomaly |

**Fig. 2.** The typology of anomalies

The second dimension is the **cardinality of relationship** and represents how the various attributes relate to each other when describing anomalous behavior. These attributes, individually or jointly, are responsible for the deviant character of the case:

- *Univariate*: Except for being part of the same set, no relationship between the variables exists to which the anomalous behavior of the deviant case can be attributed. To detect the anomaly, its attributes can therefore be analyzed separately – i.e. the analysis can assume independence between the variables.
- *Multivariate*: The deviant behavior of the anomaly lies in the relationships between its variables. The anomaly can thus not be detected by studying the individual attributes separately. Variables need to be analyzed jointly in order to take into account their relations, i.e. combinations of values. 'Relationships' should be interpreted broadly here, including (partial) correlations, interactions, collinearity, as well as associations between attributes of different data types.

The preliminary typology presented in [1, 5, 18] is summarized in the first row of Fig. 1. The typology presented in this paper is an updated and extended version. All data types are now treated separately, yielding six basic anomaly types. In addition, the terminology is updated. The new typology is depicted in Fig. 2. The types are illustrated in Fig. 3 and 4 (note: the reader might want to zoom in on a digital screen to see colors, patterns and data points in detail). Fig. 3.A, 3.B and 4.A are simulated datasets, while Fig. 4.B depicts real-world data from the Polis Administration, an official register of income data in the Netherlands [1]. The six types of anomalies, which follow naturally and objectively from the two dimensions, are described below.

**I. Extreme value anomaly**: A case with an extremely high, low or otherwise rare value for one or multiple individual numerical attributes [cf. 3, 19]. As such cases deviate w.r.t. one or more individual attributes, their anomalous nature does not rely on relationships between attributes. However, the more attributes take on an extremely high, low or rare numerical value, the more anomalous the case is. The two cases with label *Ia* in Fig. 3.A are examples, as are the *Ib* cases in Fig. 4.B. Traditional univariate statistics typically offers methods to detect this type, e.g. by using a measure of central tendency plus or minus 3 times the standard deviation or the median absolute deviation [3, 13, 17]. These cases are literally 'outliers', as they lie in an isolated region of the numerical space. However, note that this type includes rare cases [cf. 19] and that such low-density values can also be located in the middle of the value range.

**II. Rare class anomaly**: A case with an uncommon class value for one or multiple categorical variables. Cases of this type are anomalous w.r.t. one or more individual attributes, so the deviant nature of rare class anomalies does not rely on relationships between attributes. However, like Type I cases, the more attributes take on a rare class value, the more anomalous the case is. The research in [20] deals with this type of anomaly. Case *IIa* in Fig. 3.B is a rare class anomaly, being the only green data point in the set. Case *IIb* in Fig. 4.A, the only square class, is another example. The rare red and orange colors of Fig. 4.B's *IIc* points make for rare class anomalies as well.

**III. Simple mixed data anomaly**: A case that is both a Type I and Type II anomaly, i.e. with at least one extreme value and one rare class. This anomaly type deviates with regard to multiple data types. This requires deviant values for at least two attributes, each anomalous in its own right. These can thus be analyzed separately; analyzing the attributes jointly is unnecessary because the case is not anomalous in terms of a combination of values. However, similar to the other univariate anomaly types, the more attributes take on a rare value, the more anomalous the case in question is. Case *IIIa* in Fig. 4A is an example. Case *IIa* in Fig. 3.B would be a Type III anomaly if it had been positioned to the extreme left (at the location of label 'IIa').

**IV. Multidimensional numerical anomaly**: A case that does not conform to the general patterns when the relationship between multiple continuous attributes is taken into account, but which does not have extreme values for any of the individual attributes that partake in this relationship. The anomalous nature of a case of this type lies in the deviant or rare combination of its continuous attribute values, and as such hides in multidimensionality. It therefore requires several continuous attributes to be analyzed jointly to detect this type. A multidimensional numerical anomaly in independent data is literally 'outlying' with respect to the relatively dense multivariate clouds or local patterns, and is thus located in an isolated area [cf. 21]. Case *IVa* in Fig. 3.A is an example, as well as the *IVb* cases in Fig. 4.B. In dependent data the focus may lie on one substantive attribute (e.g. 'amount spent'), although at least one other attribute is still needed to link the related individual cases. See the discussion on examples 5.B, 5.C and 5.D in section 4 for more information. So-called 'contextual' [13] or 'conditional' [14] anomalies should be seen as a special case of a multi-dimensional numerical anomaly. These require that the respective contextual or environmental attributes, such as time or location, are denoted explicitly. This explicit denotation is allowed, but not demanded, for Type IV anomalies.

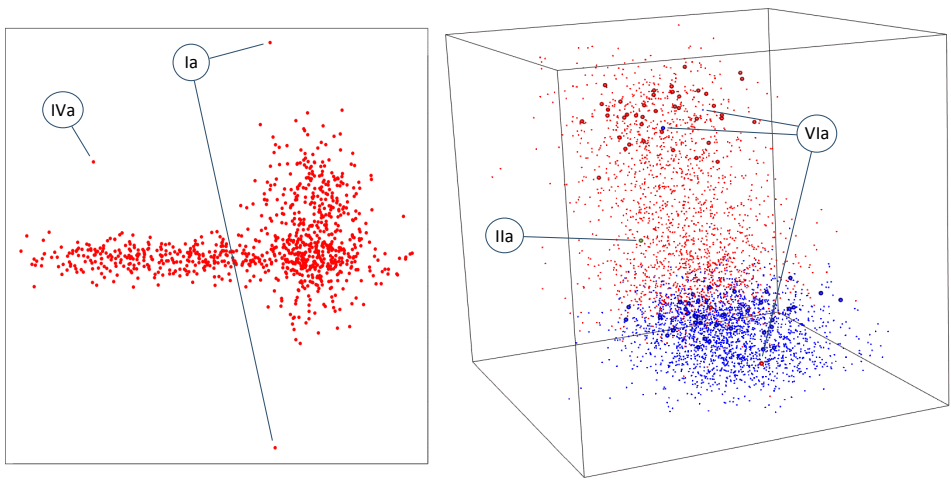

**Fig. 3.** (A) Set with two numerical variables; (B) Set with three numerical attributes and two categorical attributes (color and size)

**V. Multidimensional rare class anomaly**: A case with a rare combination of class values. In datasets with independent data points a minimum of two substantive categorical attributes needs to be analyzed jointly to discover a multidimensional rare class anomaly. An example is this curious combination of values from three attributes used to describe dogs: 'MALE', 'PUPPY' and 'PREGNANT'. Another example is case *Va* in Fig. 4.A, as it is the only red circle in the set. When dealing with dependent data (see sections 2 and 4), the anomaly can also be a deviant combination of class values of a single substantive attribute, but from multiple related cases. Again, an additional attribute, such as time, is still required to link these dependent cases. An example of such a type V anomaly in dependent data is the deviant phase-sequence in section 4.

**VI. Multidimensional mixed data anomaly**: A case with a class or a combination of classes that in itself is not rare in the dataset as a whole, but is only rare in its local pattern or neighborhood (numerical area). The anomalous nature of a case of this type lies in the deviant relationship between its continuous and categorical attributes. As with the other multivariate anomalies, such cases hide in multidimensionality and thus multiple attributes need to be jointly taken into account to identify them. As a matter of fact, multiple data types need to be used, as anomalies of this type per definition are comprised of both numerical and categorical attributes. Cases *VIa* in Fig. 3.B are illustrations of a multidimensional mixed data anomaly in independent data, as they are points with a color rarely seen in their respective neighborhoods. This also holds for the *VIb* cases in Fig. 4.B, being blue points in an otherwise pink local pattern (or vice versa). Type VI cases can also take the form of second- or higher-order anomalies, with categorical values that are not rare (not even in their neighborhood), but are rare in their combination in that specific area. Here is another way to look at this: a first-order Type VI anomaly can be seen as a *rare class anomaly* in its local neighborhood, while a second- or higher-order Type VI anomaly can be seen as a *multidimensional rare class anomaly* in its local neighborhood.

More examples of the different types of anomalies will be provided in the Discussion section. Additional illustrations (including those of higher-order anomalies) can be found in [1].

## 4    Discussion

The typology presented here offers a clear and tangible definition of the different types of anomalies. As the various figures show, these types lend themselves to be clearly illustrated by visual plots. In addition to providing a clear *understanding* of the different kinds of anomalies that exist, the typology can be used to *evaluate* AD algorithms. This is a relevant contribution because most research publications do not make it very clear which types can be detected by the anomaly detection algorithms presented, even though it is clear that many of those algorithms are incapable of identifying all types [1, 11]. It is therefore advised that researchers use the typology to provide clear insight into the functional capabilities of their AD algorithms by explicitly stating which anomaly type(s) can be detected. This also gives due acknowledgment of the *no free lunch* theorem in an AD context [cf. 10, 11, 12].

**Evaluation of algorithms.** Using the typology for algorithm evaluation has more implications than merely stating which types can be detected, since the typology is ideally also used to *create test sets*. AD studies often evaluate algorithms by treating (a sample of) one class in existing datasets as anomalies [22]. However, this is a questionable practice because these classes may actually represent true patterns rather than true deviants, and may be very similar to other classes in the dataset. This latter situation can indeed be observed for several classes in the real-world Polis dataset.

Moreover, there is no guarantee that all anomaly types will be present in such a test set. A better approach for creating AD test sets would therefore be to take the typology presented here and insert several instances of each anomaly type in a simulated or real-world dataset. This ensures that the different types of anomalies are present in the set and a thorough evaluation of the algorithm can thus be conducted. Researchers should at least aim to include the most important types, based on the domain or the problem being studied. See [1] for an example of an evaluation.

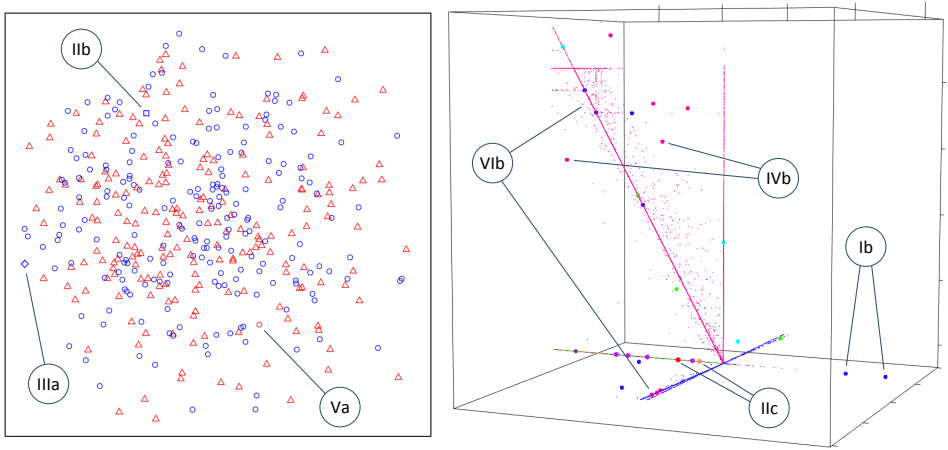

**Fig. 4.** (A) Set with two numerical attributes and two categorical attributes (color and shape); (B) Polis Administration set with one categorical and three numerical attributes, and large dots representing the 30 most extreme anomalies detected by SECODA [1]

**Local vs. global anomalies.** The typology presented here also offers a natural way to distinguish between *local* and *global* anomalies. It follows from the typology that the three univariate anomaly types are global anomalies, as these are unusual w.r.t. an individual attribute (possibly several individual attributes, but each attribute is anomalous in its own right). They are anomalous with regard to the entire dataset. When taking all the set's cases into account, extreme value anomalies will always have an extremely low, high or rare value for the given attribute. Rare class anomalies and simple mixed data anomalies likewise have an extremely rare value for the given attribute(s), without any condition and regardless of the other attributes. The three multivariate anomaly types, on the other hand, are only deviant given the categorical condition or the specific numerical area the case in question is located in. This is the result of the fact that the anomalous nature of the case lies in the combination of its attribute values. This is clearly illustrated by the *VIb* cases in Fig. 4.B, where the blue cases are not anomalous because they are blue (which is a very normal color in the set), but because they seem to be misplaced between the pink cases. A similar argument holds true for the pink *VIb* cases. In short, the three multivariate anomaly types are situational and therefore local. The three univariate anomaly types represent global anomalies, as they are anomalous regardless of the values of other attributes.

**Other typologies.** The typology presented here can be used both for *clarifying* more abstract typologies and for positioning the anomalies of more specific typologies in a *broader framework*. The typology presented in [13] is very general in nature, yielding rather abstract anomaly types. This is made clear by the fact that, given some assumptions, all six anomaly types of this paper's typology can manifest themselves as a point anomaly. A *point anomaly* is simply an individual case that deviates from the rest of the data [13]. The example is given of a very high 'amount spent' in a dataset with credit card transactions. This is exactly what the *extreme value anomaly* in the typology of Fig. 2 is. Another example in [13] concerns isolated cases that are described by two numerical dimensions, of which none has an extremely high or low value. This is therefore this paper's *multidimensional numerical anomaly*. The explanation in [13] does not explicitly state whether point anomalies can also be comprised of categorical data. However, if this type is interpreted in a broad sense, then *rare class anomalies* are also point anomalies because these are unique or rare data points and there is no need for dependent data or an explicitly denoted context (see below). A similar argument can then be made for *multidimensional rare class anomalies*, *multidimensional mixed data anomalies* and *simple mixed data anomalies*, which renders point anomalies a very broad and abstract type indeed. The typology presented in Fig. 2 is thus helpful, and even needed, to obtain a more concrete understanding of how point anomalies can manifest themselves.

The *contextual anomaly* in [13] is only deviant in a specific and explicitly specified context, such as a certain location or time period. This requires relationships between variables, making this a *multidimensional numerical anomaly* (and possibly a *multidimensional rare class anomaly* or *multidimensional mixed data anomaly*) for which the analyst has explicitly specified the contextual variables before running the analysis.

Finally, the *collective anomaly* in [13] refers to a group of cases that, as a combined whole, shows deviant behavior. An example is when individual cases are not deviant in themselves, but only as a group of cases that represents a deviant sequence. The set of red underlined classes in the following phase-sequence can therefore be regarded as such an anomaly:

phase1, phase2, phase3, phase1, phase2, phase3, <u>phase1, phase3,</u> phase1, phase2, phase3

In terms of this study's typology this is a *multidimensional rare class anomaly*, in which the combination (sequence) of classes deviates from the regular pattern (the cycle 'phase1, phase2, phase3'). As one can see from the example, the anomaly is comprised of multiple cases in a set with dependent data (related points or rows). If relevant, however, one could abstract from the original individual points and declare the anomaly at the group level (i.e. the cycle), turning this into a *rare class anomaly*.

Collective anomalies in sequence data can be described in more detail both by the typology presented in this paper and by the specific typologies from time series analysis [15, 16]. Additional examples are shown in Fig. 5 and will be discussed below. In time series analysis, the left red spike of Fig. 5.A is an *additive outlier*, the right spike a *transitory change outlier* that takes some time to disappear [15]. In terms of this paper's typology, both are *extreme value anomalies* as they have an extremely high respectively low value for a single attribute. The isolated spike in Fig. 5.B also constitutes an *additive outlier*. However, this is not an *extreme value anomaly*, as it deviates from the local pattern without exhibiting extreme values from a global perspective. This is therefore a *multidimensional numerical anomaly*, as it requires two numerical attributes to identify the anomaly in the local pattern. Interestingly, the typology of Fig. 2, albeit in principle more abstract than a specific typology dedicated to time series anomalies, is thus able to distinguish between instances of one and the same time series type. The typology's ability to make this more specific distinction is due to its fundamental dimensions: data types and cardinality of relationship.

The red transition of Fig. 5.C constitutes a *level shift outlier*. This can be regarded as a collective anomaly because no individual point is anomalous – the deviation lies in the sudden level shift of the sequence. The deviant behavior can only be detected by taking into account both the time and the value variable, making this a *multidimensional numerical anomaly*. However, by first determining the difference between two consecutive cases, this change point detection problem can be turned into a simple search for extreme values. Given the transformed dataset that results from this operation, this would thus be an *extreme value anomaly*, representing a deviant transition – which is now a single case – rather than (a group of) cases from the original set.

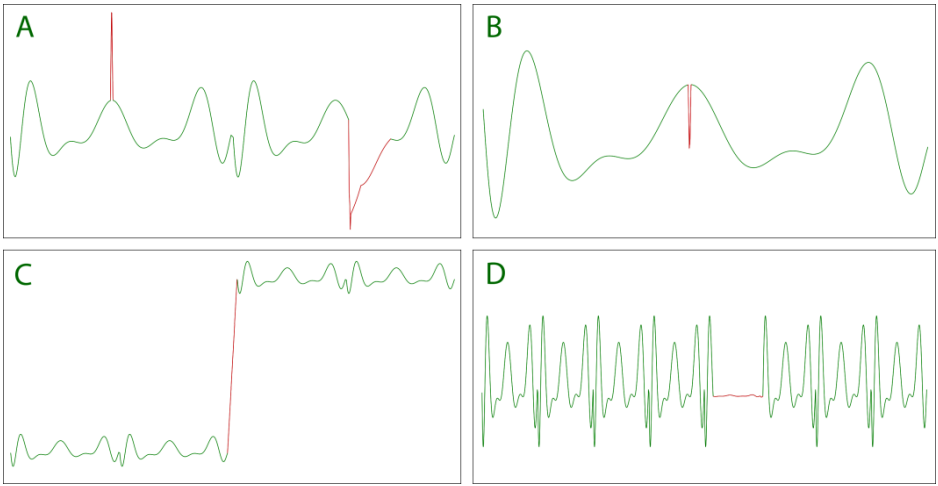

**Fig. 5.** Four time series with time on the horizontal axis and the anomalies in red

The red part of Fig. 5.D is another example of a collective anomaly, in the form of a *deviant cycle anomaly*. As this involves numerical data of which the red part – which does not feature extremely high, low or rare individual values – is found to be anomalous when taking into account the entire sequence, this is a *multidimensional* *numerical anomaly*. Re-ordering the sequence in a random fashion (i.e. discarding the time attribute) would make the anomaly 'disappear', leaving neither a *deviant cycle anomaly* nor a *multidimensional numerical anomaly.* It is similar to Fig. 5.B's anomaly, except for the fact that 5.D's anomaly represents a group (cycle) of related cases instead of a single data point. Note that collective anomalies such as those in 5.D can usually only be discovered in sets with dependent data and by specialized algorithms [cf. 13, 22]. As with 5.C, the example of 5.D can be transformed from a *multidimensional numerical anomaly* into a simpler type. The individual cycles could first be detected and classified, after which a *rare class anomaly* can be denoted.

**Conceptual levels.** The examples of 5.C, 5.D and the phase-sequence make clear that one can *abstract from* the original micro-level data points to view the data and anomalies at a higher and somewhat simpler conceptual level. The grouping variables, such as time, location and identification attributes, are generally used for this. In 5.D (and the phases example) the sequence data were also analyzed at the level of a cycle, transforming a Type IV (and Type V) anomaly into a *rare class anomaly*. In terms of the typology of Fig. 2 this implies a change from a multivariate anomaly to a univariate one. In terms of [13] this changes the anomaly from a collective to a point anomaly. In addition to the change in conceptual perspective, this may involve a different AD algorithm or a transformation of the dataset. There can be several reasons to change the conceptual level of a dataset and its anomalies. First, the goal of the analysis may imply a certain conceptual focus. The aim may be to detect anomalous individual data points (e.g. logged events such as login attempts) or aggregated entities (e.g. entire user sessions comprised of multiple actions). A second reason to change the level of a dataset is the fact that some sets may be too big to process. The data reduction obtained by transforming the dataset into a set with aggregated cases may be required to make the analysis more manageable. A third reason concerns the AD algorithms the analyst has at his or her disposal. An advanced algorithm to analyze dependent data may simply not be available, meaning that the analyst first needs to transform the dataset to a format that is suitable for the algorithms at hand.

**Terminology.** To conclude this discussion, it is worthwhile to re-assess the synonyms mentioned at the beginning of this paper. As stated, the terms *anomaly*, *outlier*, *novelty* and *deviant* are often treated as having an identical meaning. However, in light of the typology and discussion presented here, several of these terms should be defined more clearly. The term *outlier*, from a traditional statistical perspective, refers to observations that literally lie outside the general patterns or dense data clouds. In other words, such cases lie in a numerically isolated region of the space. Given this typology, the term *outlier* can thus best be reserved for *extreme value anomalies*, *simple mixed data anomalies* and, in the case of independent data, *multidimensional numerical anomalies*. Likewise, the term *novelty* can be defined more strictly, as this should refer to cases that in some way represent new and hitherto unknown events or objects. Therefore, this term can best be applied to situations in which a case represents something that has not happened or been detected before. This could be the case in change point detection analysis, such as in the time series of Fig. 5.

Alternatively, a *novelty* could refer to cases discovered with unsupervised or one-class anomaly detection, in which the identified case is not a data point from a pattern that the algorithm has learned before by training on labelled data. Finally, the terms *anomaly* and *deviant* can be regarded as general terms and true synonyms.

## 5      Conclusion

This paper has presented a general typology of anomalies that offers a concrete understanding of the different anomalies one can encounter in datasets. The typology can also be used to evaluate AD algorithms in a more transparent way. In particular, researchers can utilize it to create test sets that will be used in the evaluation and should report explicitly which types of anomalies can be detected by a given AD algorithm. Furthermore, as a result of its fundamental dimensions, the typology can be used both for clarifying existing typologies that are more abstract in nature [e.g. 13] and for studying the anomalies of specific typologies [e.g. 15] through a more general lens. For some specific, dedicated typologies, this study's typology can even provide deeper insight by proposing meaningful sub-divisions within existing types. Finally, the typology clearly distinguishes between local and global anomalies, and can be used as a framework to analyze the conceptual levels of data and anomalies.

**Remarks.** The data examples and the R code to analyze them can be downloaded from www.foorthuis.nl. The author thanks Emma Beauxis-Aussalet for her valuable remarks.